\documentclass[]{interact}
\usepackage[utf8]{inputenc}
\usepackage[T5]{fontenc}
\usepackage{scalerel,stackengine}

\usepackage{epstopdf}
\usepackage[caption=false]{subfig}

\theoremstyle{plain}

\theoremstyle{definition}

\theoremstyle{remark}

\usepackage{algpseudocode}
\usepackage{algorithm}

\usepackage{forest}
\usetikzlibrary{shadows}

\usepackage{tikz, pgfplots} 
\pgfplotsset{compat=1.18}
\usetikzlibrary{positioning}
\usetikzlibrary{calc} 
\usepackage{url}

\begin{document}


\title{Leveraging Sentence-oriented Augmentation and Transformer-Based Architecture for Vietnamese-Bahnaric Translation}


\author{
\name{Sang T. Nguyen\textsuperscript{$^\ast$$\dagger$}, Nguyen Q. Pham\textsuperscript{$^\ast$$\dagger$$\ddagger$}, Tho Quan\textsuperscript{$^\ast$$\dagger$}\thanks{CONTACT Nguyen Q. Pham Email: quocnguyenkh@gmail.com}}
\affil{\textsuperscript{$^\ast$}Falcuty of Computer Science and Engineering, Ho Chi Minh City University of Technology (HCMUT), 268 Ly Thuong Kiet, District 10, Ho Chi Minh City, Vietnam}
\affil{\textsuperscript{$\dagger$}Vietnam National University Ho Chi Minh City (VNU-HCM), Linh Trung Ward, Thu Duc District, Ho Chi Minh City, Vietnam }
\affil{\textsuperscript{$\ddagger$}Corresponding author: \texttt{quocnguyenkh@gmail.com} }
}

\maketitle

\begin{abstract}
The Bahnar people, an ethnic minority in Vietnam with a rich ancestral heritage, possess a language of immense cultural and historical significance. The government places a strong emphasis on preserving and promoting the Bahnaric language by making it accessible online and encouraging communication across generations. Recent advancements in artificial intelligence, such as \textit{Neural Machine Translation} (NMT), have brought about a transformation in translation by improving accuracy and fluency.
This, in turn, contributes to the revival of the language through educational efforts, communication, and documentation. Specifically, NMT is pivotal in enhancing accessibility for Bahnaric speakers, making information and content more readily available. Nevertheless, the translation of Vietnamese into Bahnaric faces practical challenges due to resource constraints, especially given the limited resources available for the Bahnaric language. To address this, we employ state-of-the-art techniques in NMT along with two augmentation strategies for domain-specific Vietnamese-Bahnaric translation task. Importantly, both approaches are flexible and can be used with various neural machine translation models. Additionally, they do not require complex data preprocessing steps, the training of additional systems, or the acquisition of extra data beyond the existing training parallel corpora.
\end{abstract}

\begin{keywords}
Data augmentation; low-resource neural machine translation; machine translation; Bahnar language
\end{keywords}

\section{Introduction}
The Bahnar people, sometimes called Ba-Na, represent a distinct ethnic minority in Vietnam's diverse spectrum of ethnic groups. At present, the Vietnamese government is actively striving to enable their full participation in the wider society. This effort primarily concentrates on enhancing their participation in social and cultural aspects, as well as in the fields of education and science. As a component of this program, there is a significant focus on translating crucial documents into the Bahnar language. Therefore, machine translation has been addressed as a possible solution for translating Vietnamese to Bahnar. With the emergence of deep learning in recent years, Neural Machine Translation \cite{Kalchbrenner2013RecurrentCT, Sutskever14} has become a new model and become the mainstream of machine translation. NMT helps establish an accessible means of communication for Bahnar people. This research is not merely focused on scientific techniques; it is also a means of preserving the national language and paying tribute to the spiritual values and culture of the Bahnar people.

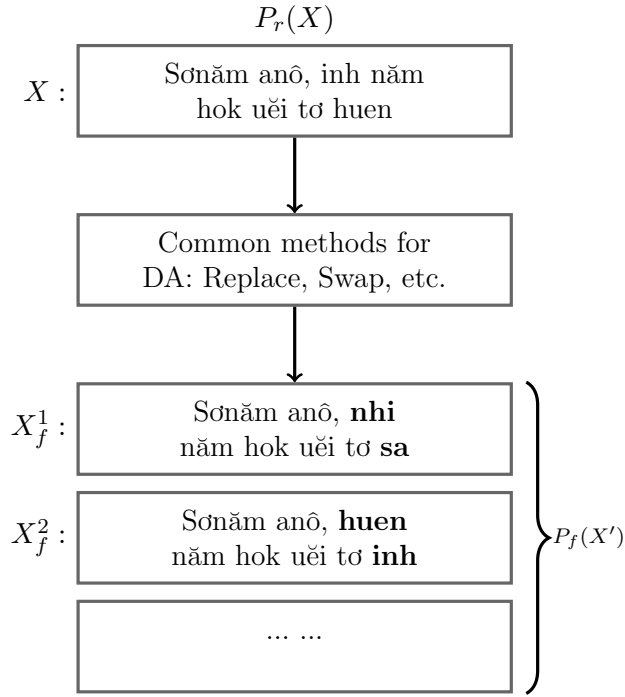
\begin{figure}[H]
\centering
    \begin{tikzpicture}[
    SIR/.style={rectangle, draw=black!60, very thick,
    text centered,
    inner sep=0pt, 
    text width=5.7cm, 
    text height=5mm, 
    text depth=7mm
    },
    ]
    \node[SIR, label=west:$X:$, label=above:$P_r(X)$] (origin) {Sơnăm anô, inh năm hok uĕi tơ huen};
    \node[SIR] (methods) [below=of origin] {Common methods for DA: Replace, Swap, etc.};
    \node[SIR, label=west:$X^1_f:$] (replace) [below=of methods] {Sơnăm anô, \textbf{nhi} năm hok uĕi tơ \textbf{sa}};
    \node[SIR, node distance=2mm, label=west:$X^2_f:$] (swap) [below=of replace] {Sơnăm anô, \textbf{huen} năm hok uĕi tơ \textbf{inh}};
    \node[SIR, node distance=2mm] (more) [below=of swap] {... ...};

    \draw[->, very thick] (origin.south) to node[right] {} (methods.north);
    \draw[->, very thick] (methods.south) to node[right] {} (replace.north);
    \draw[very thick, black, decoration={brace, raise=5pt, amplitude=3mm}, decorate] (replace.north east) -- (more.south east) node [black, midway, xshift=1cm]{\footnotesize$P_f(X')$};
    
    \end{tikzpicture}
    \caption{The commonly used methods of DA for NMT}
    \label{fig:common_da}
\end{figure}

While the availability of large parallel corpora significantly impacts how an NMT system performs, the Bahnar language itself is a low-resource language \cite{thuylinh.2018}, which can make the system suffer from poor translation quality \cite{kocmi-bojar-2018-trivial}. Therefore, \textit{Data Augmentation} (DA) \cite{lecun-95a} needs to be involved in the project to generate extra data points from the empirically observed training set to train the NMT model. With DA, the performance of the translation system can be significantly improved, all while making efficient use of the current resources. Data augmentation was first widely applied in the computer vision field and then used in natural language processing, achieving improvements in many tasks. DA helps to improve the diversity of training data, thereby helping the model anticipate the unseen factors in testing data. DA applications in natural language processing have been investigated in recent years, and the most well-known fields are text classification \cite{ wang-yang-2015-thats, wei-zou-2019-eda, Zhang2020a}, text generation (including NMT) \cite{sennrich-etal-2016-improving, wang-etal-2018-switchout, DBLP:journals/corr/ZhangZL15}, and structure prediction \cite{DBLP:journals/corr/abs-2012-11468, Longpre2020}. DA is still a super common and all-over-the-place approach in NMT, which samples some fake data distribution $P_f(X')$ using some common methods (Figure \ref{fig:common_da}) based on real data distribution $P_r(X)$, where $X_f^1$, $X_f^2$ refer to augmented data generated from real data using common approaches, such as replacing, swapping.

In this paper, researchers have explored and tested two DA methods. Both of these approaches can produce parallel sentences that, even though they are highly improbable based on the data distribution, consistently enhance the quality of the resulting NMT system. The first method follows the main idea of the method \textit{Easy Data Augmentation} (EDA) \cite{wei-zou-2019-eda}, which is a framework that contains a simple set of chosen DA techniques for natural language processing. However, the DA procedures employed in this method are inspired by the findings of multi-task learning data augmentation (MTL DA) research \cite{DBLP:journals/corr/abs-2109-03645}. As a result, researchers have chosen to name it MTL DA as well. This approach generates artificial target sentences with the purpose of reinforcing the encoder. The second approach, drawing inspiration from the work of \cite{DBLP:journals/corr/abs-2010-11132}, takes a distinct path by extending the application of the noise-based augmentation technique beyond the word level. Researchers have examined the potential of this method in different contexts beyond the original research, which primarily focused on low-resource NMT. Several extensive experiments have been conducted and delved deeply into the hyperparameters that govern the behavior of this method. 

Neither method necessitates complex preprocessed procedures, training additional systems, or utilizing data beyond the existing training parallel corpora. Experiments were conducted to assess the performance enhancement achieved by these methods compared to the previous research (EDA) and the widely recognized augmentation technique (word replacement using semantic embedding). In addition to evaluating their effectiveness on the entire dataset, both methods have been specifically researched and employed to address certain sentence forms that pose significant challenges in the translation task.

Contributions of our work are as follows:
\begin{itemize}
\item Studying and implementing two data augmentation techniques, including multi-task learning data augmentation and sentence boundary augmentation, to enhance low-resource language data, thus improving the performance of the NMT model.
\item A more in-depth study was conducted on each data augmentation method by selecting the most suitable data augmentation operations and parameters for the Vietnamese to Bahnar translation task.
\item Effectively applying the mentioned data augmentation methods helps in improving common errors (wrong collocation and word-by-word) in Vietnamese-Bahnar translation.

\end{itemize}

The remainder of the paper is organized as follows. The next section briefly provides background knowledge of neural machine translation, data augmentation, and its primary techniques. After that, Section \ref{sec:bana} gives an overview of the Bhanar language and the differences between Vietnamese and Bahnar. Section \ref{sec:relatedworks} presents a review of the most relevant works in the area of DA for NMT. The details of two methods MTL DA and sentence boundary augmentation, are shown in Section \ref{sec:approaches}, whereas Section \ref{sec:experi} describes the experimental settings and discusses the results obtained. Finally, the paper ends with some concluding remarks in Section \ref{sec:conclu}.

\section{Preliminaries}
\subsection{Neural machine translation}
\subsubsection{Overview}
\textit{Machine Translation} (MT) \cite{Budiansky1951} is a major sub-field of \textit{Natural Language Processing} (NLP) \cite{Jurafsky2009} that focuses on translating human languages automatically by using a computer. In the early stages, machine translation relies heavily on manual translation rules and linguistic knowledge. However, because the nature of language is significantly complicated, it is impossible to cover all irregular cases with just hand-crafted translation rules. During the development process of MT, more and more large-scale parallel corpora appeared. With the data-driven approaches, \textit{Statistical Machine Translation} (SMT) \cite{Brown1990ASA} has replaced the original rule-based translation due to its availability to study latent factors such as word alignment or phrases directly from corpora. However, SMT is still far from expectations because it is unable to model long-distance word dependencies. With the emergence of deep learning, Neural Machine Translation has become a new model and replaced SMT to become the mainstream of MT.

\subsubsection{Modeling}
Assuming a source sentence x = \{$x_{1}$,..., $x_{S}$\} and a target sentence y = \{$y_{1}$,..., $y_{T}$\} are given. By using the chain rule, the conditional distribution of a standard NMT \cite{DBLP:journals/corr/BahdanauCB14, DBLP:journals/corr/VaswaniSPUJGKP17} can factorize a sentence-level translation probability as a product of word-level probabilities from left-to-right (L2R) as: 

\begin{equation} \label{eq:nmt_overview}
P(\textbf{y}|\textbf{x}) = {\displaystyle \prod_{t=1}^{T} P(y_{t}|y_{0},..y_{t-1}, x)}
\end{equation}

NMT models that conform to Eq. \ref{eq:nmt_overview} are referred to as L2R autoregressive NMT \cite{Kalchbrenner2013RecurrentCT, Sutskever14} for the prediction at time-step \textit{t} is taken as input at time-step \textit{t} + 1.

NMT normally uses \textit{maximum log-likelihood} (MLE) as the training objective function, which is usually used for estimating the parameters of a probability distribution. Given the training corpus $\mathcal{D} = \{\langle x^{(s)}, y^{(s)} \rangle\}_{s=1}^{S}$, the goal of training is to find a set of model parameters that maximize the log-likelihood on the training set:

\begin{equation}
    \hat{\theta}_{MLE} = \underset{x}{argmax}\{\mathcal{L}(\theta)\},
\end{equation}
where the log-likelihood is defined as
\begin{equation}
    \mathcal{L}(\theta) = \sum_{s=1}log P(\textbf{y}^{(s)}|\textbf{x}^{(s)};\theta)
\end{equation}

By the back-propagation algorithm, the gradient of $\mathcal{L}$ can be computed with respect to $\theta$. NMT model training usually adopts the \textit{stochastic gradient search} (SGD) algorithm. Instead of computing gradients on the full training set, SGD computes the loss function and gradients on a mini-batch of the training set. The plain SGD
optimizer updates the parameters of an NMT model with the following rule:
\begin{equation}
    \theta \leftarrow \theta - \alpha \bigtriangledown \mathcal{L}(\theta),
\end{equation}
Where $\alpha$ is the learning rate. The parameters of NMT are guaranteed to converge into a local optima with a well-adjusted learning rate. In reality, adaptive learning rate optimizers like Adam \cite{adam} are found to significantly reduce training time compared to basic SGD optimizer.

\subsection{Data augmentation} \label{section:da}
\subsubsection{Overview}
Data augmentation refers to techniques for increasing training data diversity without collecting extra data. Most methods either produce synthetic data or add slightly modified versions of existing data, expecting that the augmented data can serve as a regularizer and lessen overfitting while training machine learning models \cite{DBLP:journals/corr/abs-1806-03852, Khoshgoftaar2019}. DA has been widely employed in computer vision, where model training typically includes operations like cropping, flipping, and color transforming. In NLP, where the input space is discrete, it is less evident how to create efficient augmented instances that capture the desired invariances.

\subsubsection{Goals and trade-offs}
Many DA strategies for NLP, from rule-based manipulations \cite{DBLP:journals/corr/ZhangZL15} to more complex generative systems \cite{liu-etal-2020-data}, have been developed despite challenges associated with the text. Since the goal of DA is to offer an alternative for collecting more data, the optimal DA technique should be simple to use while also enhancing model performance. Most offer trade-offs between these two.

Rule-based methods are simple to apply but typically result in small enhancements in performance \cite{wei-zou-2019-eda, Xu2021MDAMD}. Conversely, approaches that utilize trained models may require more resources to implement but introduce greater data variability, resulting in more substantial performance improvements. Tailored model-based techniques for specific tasks can significantly impact performance but can be challenging to develop and utilize effectively.

Additionally, it is important that the augmented data distribution strike a balance between being neither too similar nor too different from the original data. If the augmented data is too similar, it may result in overfitting. At the same time, if it is too different, it can lead to poor performance due to training on examples that do not accurately represent the intended domain. Therefore, effective data augmentation approaches should strive for a harmonious equilibrium.

Discussing the interpretation of DA, \cite{DBLP:journals/corr/abs-1803-06084} note that "\textit{data augmentation is typically performed in an ad-hoc manner with little understanding of the underlying theoretical principles}", and claim that it is insufficient to explain DA as \textit{regularization}. In general, there is a noticeable absence of comprehensive research on the precise mechanisms underlying the effectiveness of DA. Existing studies mostly focus on superficial aspects and seldom delve into the theoretical foundations and principles involved.

\subsubsection{Techniques} \label{subsection:da_techniques}
The general approaches of DA techniques are mentioned in the survey of \cite{feng-etal-2021-survey}, the DA techniques can be grouped as rule-based techniques, interpolation techniques, and model-based techniques. For rule-based techniques, these techniques do not require model components and use simple, preset transforms. A typical example of these methods is EDA \cite{wei-zou-2019-eda}, which performs a set of random perturbation operations on the token level such as random insertion, swap, and deletion. A different category of DA techniques, called interpolation, initially introduced by MIXUP \cite{mixupZhang18}, involves interpolating the inputs and labels from multiple real examples. With the model-based approach, these DA techniques from this approach have also utilized Seq2seq models and language models. An example is the widely used "back-translation" approach \cite{sennrich-etal-2016-improving}, which involves translating a sequence into a different language and then translating it back to the original language.

However, DA approaches can be more specifically categorized based on the characteristics of the techniques and the diversity of augmented data. \cite{DBLP:journals/corr/abs-2110-01852} frame DA methods into three categories, including paraphrasing, noising, and sampling.
 
\begin{itemize}
    \item The paraphrasing-based methods generate augmented data that retains a strong semantic resemblance to the original data by making controlled and limited modifications to the sentences. The augmented data effectively conveys almost identical information as the original data.
    \item The nosing-based methods aim to enhance the model's robustness by introducing discrete or continuous noise while ensuring the validity of the data. These methods focus on adding noise in a controlled manner to improve the model's ability to handle different scenarios.
    \item The sampling-based methods excel at understanding the data distributions and generating novel data samples from within these distributions. By employing artificial heuristics and trained models, these techniques produce more diverse data that effectively caters to a wider range of requirements for downstream tasks.
\end{itemize}

\subsubsection{Method stacking}
The methods described in Section \ref{subsection:da_techniques} are not restricted to being used independently. They can be combined to achieve improved performance. Some common combinations include:
\begin{itemize}
    \item \textbf{The Same Type of Methods}: Certain studies integrate various approaches based on paraphrasing and generate diverse paraphrases to enhance the diversity of augmented data. For example, \cite{LIU2020105918} uses both thesaurus and semantic embedding. Regarding methods based on noising, they often combine different techniques that were previously considered unlearnable, as demonstrated in \cite{DBLP:journals/corr/abs-2004-13952}. This is because these methods are straightforward, efficient, and mutually beneficial. Some methods also adopt different sources of noising or paraphrasing like \cite{xie2017data}. The combination of different resources could also improve the robustness of the model
    \item \textbf{Unsupervised Methods}: In certain situations, there is a need for straightforward and task-agnostic unsupervised data augmentation methods. Consequently, these methods are grouped together and extensively utilized. EDA is a very popular method that consists of synonym replacement, random insertion, random swap, and random deletion.
    \item \textbf{Multi-granularity}: Certain studies employ the same approach at various levels to enhance the augmented data by introducing diverse changes of varying granularities. This practice aims to enhance the model's robustness. For example, \cite{wang-yang-2015-thats} trains both word embeddings and frame embeddings by Word2Vec.
\end{itemize}

\newcommand\bbar{\ThisStyle{\ensurestackMath{%
  \stackengine{-.1\LMpt}{\SavedStyle\mathsf{b}}{\SavedStyle\bar{}%
  \mkern3.5mu}{O}{r}{F}{F}{L}}}}
  
\section{Bahnar language} \label{sec:bana}

\subsection{Overview} \label{sec:banaoverview}
Bahnar language or Ba-na language is a Mon-Khmer language belonging to the Bahnaric group and used by Bahnar people mainly living in central Vietnam \cite{sotayphuongngubana}. Bahnaric can be divided into two sub-group: Northern Bahnar(Xêđăng, Halăng, Jeh,...) and Southern Bahnaric(Kơho, Mnông, Chrau Jro,...). Bahnar language is an intermediate language of these two groups. Bahnar language is similar to Southern Bahnaric. Besides, the structure of its phonemes is simpler than the Northern Bahnaric. However, it shares more standard features in the vocabulary with Northern Bahnaric. Therefore, the Bahnar language can be classified as Central Bahnaric.

Vietnamese-Bahnar has unique differences from other translation pairs. From the perspective of Vietnamese-English translation, grammar plays a major role that requires the conversion of words into different variants depending on tense, active or passive voice, and in some specific cases, words or phrases may express different meanings which is the cultural characteristic of the language, slang for example. In terms of translation into languages containing pictograms (Chinese), information-rich and diverse word meanings require a high ratio of purposeful compression while preserving the correctness of the mentioned entities. In Vietnamese-Bahnar translation, in addition to the above problems, although there are similarities in sentence structure between the two languages, the existence of many sub-syllables with various accents, rare sub-syllables, and cultural slang result in challenges for learners and also machine translation algorithms. 

To be more specific, in terms of complexion, the word structure of the Bahnar language has several solid rules that connect firmly with lexemes. Words in the Bahnar language can be constructed using "affix", "reduplication", and "compound" \cite{bahnartodinhnghia}. Method affix is the most complicated one among the three ways. These methods can create derivational words with typical features such as variation word meanings and changing grammar functionality (e.g., nouns can become verbs).

\subsection{Grammar rules}

\subsubsection{Bahnaric Vocabulary} \label{sec:bahnavocab}

Firstly, according to the Bahnaric language training program \cite{gov.2020}, the Bahnar language has a different character set than Vietnamese, containing 42 characters in total, with various accents. The above 42 characters form multiple-word components: 
\begin{itemize}
    \item 42 characters: {\it a, ă, â, b, \bbar, \v{c} d, đ, e, \v{e}, ê, \v{ê}, g, h, i, \v{i}, j, k, l, m, \~{n}, o, \v{o}, ô, \v{ô}, ơ, \v{ơ}, p, r, s, t, u, \v{u}, ư, \v{ư}, w, y, f, q, v, x, z}.
    \item 12 common diphthongs: {\it ia, iă, ie, i\v{e}, iô, i\v{ô}, ua, uă, ue, u\v{e}, uê, uê}.
    \item 52 common consonants: {\it bl, br, by, \v{c}h, \v{c}r, dj, djr, dr, gl, gr, gry, gy, hl, hm, hml, hmr, hmy, hn, hng, h\~{n}, hr, hy, jr, kh, khy, kl, kr, ky, ly, ml, mr, ny, my, \~{n}r, ng, ngl, ngr, ‘ngr, nhr, ngy, ny, ph, phr, phy, pl, pr, py, sr, th, thy, tr, ty }. 
    \item Some of these consonants are scarcely used in real-world sentences, which is a challenge for Bahnaric learners: {\it by, ly, ky...}
\end{itemize}

Another considered syntactic point is the use of {\it "'"} before consonants and pre-syllable (sub-syllable or weak syllable, which can be considered sesquisyllabic) to construct different word forms but unchanged in meanings. Some typical pre-syllables can be listed as a, bơ, dơ, hơ, jơ, etc (Example: ame (chăm nom), bơbah (cuối nguồn), hơhoi (không có), jơnap (đầy đủ), etc). 

From the view point of word form, similar to Vietnamese, Bahnaric words have single-word and compound forms, consisting of complex forms (words containing two or more syllables, each syllable is a meaningful single word), alliterative expressions such as adal adal (khe khẽ), kueng kueng (ầm ầm) or ring ring (nhộn nhịp). 

Additionally, Bahnaric words can express different meanings in daily communication situations. For example, the word {\it "adal adal"} may mean soft when talking about voice or quietly making some actions, this word also means slowly and carefully or sometimes unhurriedly. As mentioned in \ref{sec:banaoverview}, region and culture can affect vocabulary, which forms regional synonyms, resulting in suffering when mapping Bahnaric from one region to another. 

Last but not least, Bahnaric vocabulary also contains loan words (mostly from Vietnamese), constructed by removing accents from Vietnamese words. Words in this vocabulary set are practically entities, especially named entities of persons, locations, organizations, and a small set of nouns and verbs.  

\subsubsection{Sentence Structure}
The sentence structure of Bhanar is similar to that of Vietnamese. An ordinary simple sentence consists of two main components \textbf{Subject} and \textbf{Predicate}. The order of subject and predicate is the same as in Vietnamese.
    \begin{center}
    \textbf{Subject} // \textbf{Predicate}   
    \end{center}

More complex sentences can be formed by combining various simple subject-predicate structures with conjunction (nhưng - mă lĕi - but, và - sơm - and, vì... nên... - yuô... kơna... - because... then..., ...). In real-life usage, the subject part in sub-sentences may be reduced or replaced by pronouns, which leads to a co-reference resolution problem to fully understand the whole sentence and make the correct translation.

\subsection{Vietnamese-Bahnar translating notices} \label{sec:vb_tran}

Translating from one language to another requires careful consideration of numerous factors that significantly influence the accuracy and quality of the translation. 

These factors hold true when undertaking the translation process from Vietnamese to Bahnar as well, demanding ken attention and careful handling.

\begin{itemize}
    \item Spelling: A spelling error refers to a deviation from the standard or accepted way of spelling a word. While it is not a major issue, it could occur due to errors in the input files.
    \item Collocation: Concerning the question of whether a specific phrase consists of words that naturally occur together or co-occur.
    \item Grammatical: Grammatical error refers to an occurrence of incorrect, unconventional, or disputed language usage, such as a misplaced modifier or an inappropriate verb tense.
    \item Typo: A typographical error, commonly known as a typo, refers to a mistake made during the typing or printing of written or electronic material. The most common typographical error comes from Bahnaric's unique and diverse accents along with the combination of pre-syllables and {\it "'"} characters. 
    \item Word-by-word translation: Word-for-word translation is commonly understood as the process of translating text from one language to another by directly using the exact words from the original text. This issue can create grammatical issues and collocation issues because of the complexity of regional word form and daily communication circumstances [\ref{sec:bahnavocab}].
\end{itemize}

Besides, in normal conversation, there exist some different points between Bhanar and Vietnamese.

\begin{itemize}
    \item Some sentences in Bahnar tend to skip words in simple sentences. For example, "Ở Vịnh Thạnh là đông nhất" in Vietnamese can be translated to "Uei Vinh Thanh lư loi"; in this case, the word "là" in Vietnamese can be skipped during translating.
    \item The position of exclamation in Bhanar is also different from Vietnamese. In Vietnamese, exclamation words usually stay behind the predicate, but it is the opposite in Bahnar. For example, "Mẹ ơi" in Vietnamese will be translated to "Ơ Mi" in Bahnar.
\end{itemize}

\section{Related works} \label{sec:relatedworks}

\subsection{Data augmentation in NMT}

\subsubsection{Overview}
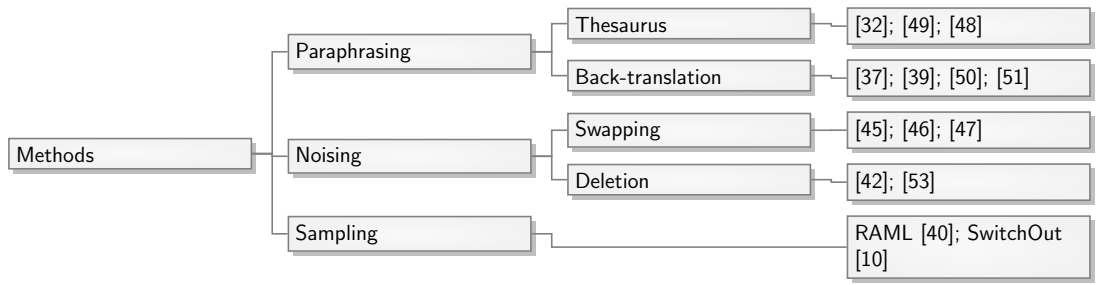
\begin{figure}[H]
    \centering
        \resizebox{\textwidth}{!}{%
        \begin{forest}
            for tree={%
                my node,
                l sep+=5pt,
                grow'=east,
                edge={gray, thick},
                parent anchor=east,
                child anchor=west,
                if n children=0{tier=last}{},
                edge path={
                    \noexpand\path [draw, \forestoption{edge}] (!u.parent anchor) -- +(10pt,0) |- (.child anchor)\forestoption{edge label};
                },
                if={isodd(n_children())}{
                    for children={
                        if={equal(n,(n_children("!u")+1)/2)}{calign with current}{}
                    }
                }{}
            }
            [Methods
                [Paraphrasing
                    [Thesaurus
                        [\cite{xie2017data}; \cite{fadaee-etal-2017-data}; \cite{gao-etal-2019-soft}]
                    ]
                    [Back-translation
                        [\cite{sennrich-etal-2016-edinburgh}; \cite{zhang-zong-2016-exploiting}; \cite{edunov-etal-2018-understanding}; \cite{caswell-etal-2019-tagged}]
                    ]
                ]
                [Noising
                    [Swapping
                        [\cite{artetxe-etal-2018-unsupervised}; \cite{lample-etal-2018-phrase}; \cite{Zhang2019RegularizingNM}]
                    ]
                    [Deletion
                        [\cite{zhang-etal-2020-token}; \cite{DBLP:journals/corr/abs-2009-13818}]
                    ]
                ]
                [Sampling
                    [RAML \cite{raml};
                    SwitchOut \cite{wang-etal-2018-switchout}
                    ]
                ]
            ]
        \end{forest}
    }%
    \caption{Taxonomy of DA methods}
    \label{fig:da_hy}
\end{figure}

In the survey of \cite{DBLP:journals/corr/abs-2110-01852}, the utilization of DA in these tasks has witnessed a notable increase in recent years. Text classification, being the pioneering task to adopt DA, has garnered a larger number of research papers compared to the other two tasks: text generation and structure prediction. In Section \ref{section:da}, it has been noted that each specific DA method has the potential to be implemented in text classification tasks. DA methods, which apply to text classification, can also apply to neural machine translation. However, due to the different nature of these tasks, some methods, which have shown powerful improvements in text classification, cannot perform well in neural machine translation. EDA is an example of the above statement. This method has created the abnormal in the context of sentences, such as: producing new vocabulary, changing word order, and skipping words. EDA will be presented in the next section and analyzed further in Section \ref{section:result_discussion} to prove its effectiveness. So, DA in NMT needs suitable approaches that are still based on the foundation of the original DA methods but need novelty modifications. Section \ref{subsection:related_da_aproach} will state the recent DA methods applied to solve this problem. The methods will be categorized according to their characteristics. Figure \ref{fig:da_hy} categorizes the typical methods in this section according to their respective DA approaches.

\subsubsection{Related works} \label{subsection:related_da_aproach}
Numerous strategies for NLP, ranging from rule-based manipulations \cite{Zhang2015CharacterlevelCN} to sophisticated generative systems \cite{liu-etal-2020-data}, have been devised despite the challenges associated with working with text.

The back-translation approach, which was introduced by \cite{sennrich-etal-2016-edinburgh}, is widely recognized as a popular DA method for NMT. However, the approaches discussed in this section primarily concentrate on methods that do not rely on additional resources apart from the available training parallel corpus.

\cite{li-etal-2019-understanding} conducted an evaluation of back-translation and forward translation in a similar context. They trained NMT systems in both forward and backward directions using the existing parallel data and then utilized these models to generate synthetic samples by translating either the target side (following the approach of \cite{sennrich-etal-2016-edinburgh}) or the source side (following the approach of \cite{zhang-zong-2016-exploiting}) of the original training corpus.

The two approaches are Reward Augmented Maximum Likelihood (RAML) \cite{raml} and its extension to the source language called SwitchOut \cite{wang-etal-2018-switchout}. These methods aim to expand the support of the empirical data distribution while maintaining its smoothness, ensuring that similar sentence pairs have similar probabilities. They achieve this by replacing words with other words sampled from a uniform distribution over the vocabulary. This approach tends to overstate infrequent words in practice. Additionally, \cite{guo-etal-2020-sequence} proposed a related approach that promotes constituent behavior, where replaced words are selected from another sentence rather than from the vocabulary.

Some auxiliary tasks have been previously used for DA, but mostly on the source side and rarely within a multi-task learning (MTL) framework. For instance, \cite{zhang-etal-2020-token} applied the technique of replacing tokens with placeholders, specifically in the source language. They combined this with auxiliary tasks that involved detecting replaced and dropped tokens. Similarly, \cite{xie2017data} evaluated the impact of replacements on the target data but did not follow an MTL approach. Another related approach is word dropout, which has been explored by \cite{10.5555/3157096.3157211} and \cite{Arora2017ASB}.

In terms of altering word order, there have been several proposals. \cite{artetxe-etal-2018-unsupervised} and \cite{lample-etal-2018-phrase} have put forward their respective strategies. However, it is notable to mention the approach suggested by \cite{Zhang2019RegularizingNM}, which involves a self-translation technique utilizing a right-to-left decoder. Their method requires generating translations from the model during training and making adjustments to multiple terms in the training loss.

There are additional noteworthy DA approaches that involve word replacement. \cite{xie2017data} employ random word replacement on the source side of training samples. \cite{gao-etal-2019-soft} replace randomly selected words with soft words, whose representations are derived from the probability distribution provided by a language model. \cite{wei-zou-2019-eda} has also applied this approach in EDA where they randomly replaced n words with their synonym. \cite{fadaee-etal-2017-data} replace several words in their training samples with infrequent words to enhance the NMT model’s performance when translating such words. They identify words to be replaced using a large source language model and then use a word-alignment model and a probabilistic dictionary to replace the corresponding counterpart with the most probable translation of the new source word.

In the context of back-translation, \cite{edunov-etal-2018-understanding} experimented with various straightforward transformations such as word deletion, replacement, and swapping on the back-translated data, resulting in a noticeable improvement. Regarding the special token used to prevent negative transfer between tasks, \cite{caswell-etal-2019-tagged} propose a similar approach to identify synthetic samples when combining actual parallel data and back-translated data for training. \cite{YANG2019240} expand upon this idea by incorporating forward-translated data for training and utilizing two distinct special tokens to distinguish between the two types of synthetic data.

\tikzset{
    my node/.style={
        draw=gray,
        inner color=gray!5,
        outer color=gray!10,
        thick,
        text width = 4cm,
        font=\sffamily,
        drop shadow,
    }
}

\subsection{Discussion}

According to empirical research, back-translation and word replacement are the two most common DA techniques for NMT. For back-translation, this technique uses monolingual data to augment a parallel training corpus. Although useful, back-translation is frequently susceptible to mistakes in initial models, a typical issue with self-training algorithms \cite{4787647_backtranslation_related}. The second category is based on word replacements. This approach is a feasible option for the Bahnar language because it produces reasonably high-quality augmented data and works best with low-resource datasets. Moreover, the noising method applied to the monolingual corpus has shown its effect with the encoder-decoder NMT model. The noising-based methods may not be the dominant approach in NMT; however, they have been researched and applied in low-level architecture during the training process. Although they have proved their potential, none of them have ever applied in NMT DA to produce visible results from augmented data like text classification. Inspired by EDA and the multi-task learning approach \cite{DBLP:journals/corr/abs-2109-03645}, a \textit{multi-task learning framework} with multiple operations should be conducted and examined to assess its effectiveness with individual methods. Besides, a further approach to the nosing-based method at the sentence level should be investigated. Therefore, \textit{combining truncated sentences on both the source and target sites beyond the word level} could provide a promising way of applying the noising-based method to low-resource NMT.

\section{Approaches} \label{sec:approaches}
To develop dependable NMT systems, a substantial volume of parallel sentences is required. These parallel sentences consist of pairs of sentences in two languages that are translations of each other. However, this poses a significant challenge for language pairs with limited resources. Building an NMT system for translating Vietnamese-Bahnar is a representative example of this challenge due to the extremely low-resource nature of Bahnar materials. Therefore, data augmentation techniques are involved in generating additional training samples when the available parallel data are scarce. To utilize the limited resources, the researcher applied some simple rule-based approaches to improve translation quality by augmenting the data. Two methods were proposed:
\begin{itemize}
    \item Multi-task Learning Data Augmentation (MTL DA)
    \item Sentence Boundary Augmentation
\end{itemize}

\subsection{Multi-task learning data augmentation (MTL DA)} \label{subsec:mtlda}

This framework was inspired by EDA \cite{wei-zou-2019-eda} and utilizes the DA operations that refer from \cite{DBLP:journals/corr/abs-2109-03645}. The approach combines a set of simple DA methods to produce synthetic target sentences to strengthen the encoder. This framework is a set of word-level augmentation methods. These methods introduce the network to novel scenarios during training where relying solely on the target language context is insufficient to achieve a minimal loss. Consequently, the responsibility is shifted to the encoder. Recent findings by \cite{voita-etal-2021-analyzing} further support this approach: they argue that the influence of source tokens in the output predictions of an NMT system decreases as decoding progresses. Moreover, this approach follows simple multi-task learning that does not require changes in the model architecture. MTL DA highlights that augmented data do not follow the distribution of a parallel corpus but can still produce a positive effect.

This framework does not require preprocessing steps, training additional systems, or data besides the available training parallel corpora, which makes it suitable for the low-resource target language, Bahnar. The framework includes five auxiliary tasks. For each task, a synthetic corpus of the same size as the original training data, obtained by transforming each original pair of sentences, will be appended to the training set. In almost all the tasks, the source sentence is left unchanged, while the target sentence is substantially modified.

Some brief explanations of each auxiliary task are presented below. Certain transformations are regulated by a hyperparameter $\alpha$, which determines the percentage of target words influenced by the transformation. Moving forward, $t$ represents the count of words in the initial target sentence. Table \ref{tab:example_mtl_da} shows an example of the effect of different methods on a single sentence pair (hyperparameter $\alpha$ is set to 0.5 for three methods: Swap, token, and replace).

\begin{itemize}
    \item \textbf{Swap:} Pairs of random target words are swapped until only $(1-\alpha)\cdot t$ remains in their original position. The objective of this task is to encourage the system to rely less on the target prefix when generating a new word.
    \item \textbf{Token:} Replacing $\alpha\cdot t$ random target words with a special (UNK) token \cite{xie2017data}. With the same motivation as \textbf{Swap}, when generating a new word, the target prefix should become less informative and force the system to pay more attention to the encoder.
    \item \textbf{Source:} The target sentence is transformed into an identical copy of the source sentence. As a result, the optimal strategy for generating the correct output is to rely on the encoder representation and replicate the relevant information from the source.
    \item \textbf{Reverse:} The order of the words in the target sentence is reversed. According to \cite{voita-etal-2021-analyzing}, the impact of the encoder diminishes as the target sentence progresses. Hence, by reversing the sentence order, the system is expected to learn to rely more on the encoder for generating words that typically occur towards the end of the sentence, utilizing additional information.
    \item \textbf{Replace:} $\alpha\cdot t$ source-target aligned words are selected at random and replaced by random words in the dictionary. Unlike \cite{fadaee-etal-2017-data} followed constrained replacements to produce only fluent target sentences, this modification is expected to introduce challenging words that cannot be easily generated solely based on the target language prefix. As a result, the system is compelled to focus on the source words and pay closer attention to them.
\end{itemize}

\begin{table}[H]
\tbl{Example of Multi-task Learning Data Augmentation.}
{\begin{tabular}{lcc} \toprule
Task & Language & Augmented sample\\ \midrule
 Original training sample & source & Bố Điêu bị ốm nặng\\ 
                            & target & Bă đe Diêu jĭ adrin \\ \midrule
    Swap & target & \textbf{Diêu} đe \textbf{Bă} jĭ adrin \\ \midrule
    Token & target & Bă đe \textbf{UNK} jĭ \textbf{UNK} \\ \midrule
    Source & target & Bố Điêu bị ốm nặng \\ \midrule
    Reverse & target & adrin jĭ Diêu đe Bă \\ \midrule
Replace & source & \textbf{con vẹt} Điêu \textbf{vàng} ốm nặng \\
            & target &  \textbf{sem diê} \textbf{'brơu} Diêu jĭ adrin \\ \bottomrule
\end{tabular}}
\label{tab:example_mtl_da}
\end{table}


The operations above will be strategically combined based on the classic stacking method of data augmentation. This approach aims to achieve optimal results with carefully selected augmented datasets. In contrast to the EDA method, which incorporates all four of its operations within a unified framework, our methodology does not aggregate all operations indiscriminately. Instead, our focus is on the thoughtful selection and integration of the most appropriate methods, guided by empirical experiments and evaluations. This principle has been substantiated in Section \ref{subsection:mtlda_exp}.

\subsection{Sentence boundary augmentation} \label{subsec:senbouda}

In this approach, the very original key idea is applying sentence-level augmentation to low-resource NMT. This approach was originally motivated by the proposal of \cite{DBLP:journals/corr/abs-2010-11132}. Their research has proved that sentence boundary augmentation has successfully overcome the issue of wrong segmentation for translating sentences produced by automatic speech recognition (ASR) systems. Moreover, \cite{Gangi2019RobustNM, zhang-zhang-2020-dynamic} have also indicated that translation degradation is caused by poor system sentence boundary prediction. It is reasonable to assume that sentence boundary errors could impact translation quality. Incorrect boundaries might separate words from crucial contextual information necessary for accurate translation. When provided with sufficient context, the system could make mistakes simply due to unexpected placements of sentence boundaries. This method effectively addresses the problem of incorrect segmentation present in the limited resources of the dataset on the Bahnar language side. Besides, this method was supported by the research of \cite{DBLP:journals/corr/abs-1909-11241}. \cite{DBLP:journals/corr/abs-1909-11241} has stated a significant point of augmenting data on the sentence level, although this research is in a different context. The researcher applied the noising-based method at the sentence level in the context of sentiment analysis for Spanish tweets. Although divided tweets were split into two halves and combined randomly sampled first and second halves with the same label, they could still keep the semantic meaning. Therefore, sentence boundary augmentation has been applied in the context of low-resource translation, and it has also exposed the model to bad segmentation during training.

\begin{algorithm}
    \caption{Sentence Boundary Augmentation \cite{DBLP:journals/corr/abs-2010-11132}}\label{alg:sentence_boundary_augmentation}
    \hspace*{\algorithmicindent} \textbf{Input}: List of source $S$, target $T$ as tuples $L= \{(S_i, T_i)\}_{i=0}^n$, hyper-parameter $p = 0.3$ for sentence length truncation
    \begin{algorithmic}[1]
        \State $L_{out} \gets list();$
        \For {$i=0,2,4,\ldots,n-2,n$}
            \State $p \sim Uniform(0, p);$
            \State $(S_1, T_1), (S_2, T_2) = L[i], L[i+1]$ 
            \State $p^{S_1}_{1}$, $p^{T_1}_{1}$ = $\lceil p \times len(S_1) \rceil, \lceil p \times len(T_1) \rceil;$
            \State $p^{S_2}_{2}$, $p^{T_2}_{2}$ = $\lceil p \times len(S_2) \rceil, \lceil p \times len(T_2) \rceil;$
            \State $S_{out} = concat([S_1[p^{S_1}_{1} : ], S_2[: p^{S_2}_{2}]]);$
            \State $T_{out} = concat([T_1[p^{T_1}_{1} : ], T_2[: p^{T_2}_{2}]]);$
            \State $L_{out}.append([S_{out}, T_{out}]);$
        \EndFor
        \State \Return $L_{out}$
    \end{algorithmic} 
\end{algorithm}

The pseudocode for this approach is given in Algorithm \ref{alg:sentence_boundary_augmentation}, which represents sentences as sequences of tokens. First, the list of pair sentences is fed to the algorithm as input. In the initial step (line 1), the algorithm initializes an empty list $L_{out}$ to store the augmented sentence pairs. In the looping process from line 2 to line 10, every two continuous sentence pairs are processed to produce a new augmented sentence pair. In line 3, portion $p$ will follow a uniform distribution of 0 and a defined hyperparameter $p$ ($p$ = 30\% by default). $(S_1, T_1), (S_2, T_2)$ are two continuous sentence pairs where $S_1, T_1$ are the first source and target sentence, and $S_2, T_2$ are the next second source and target sentence (line 4). Next, all four partitions are determined based on the ceiling value of the multiplication between the length of each respective sentence and portion $p$. The two values $p^{S_1}_{1}$, $p^{T_1}_{1}$ are the partitions of the first source sentence and the first target sentence, respectively (line 5). In line 6, $p^{S_2}_{2}$, $p^{T_2}_{2}$ are calculated similarly to previous partition values, but only with the second sentence pair. Adjacent source sentences $S$, and their corresponding target sentences $T$ are concatenated. Next, the start and end of the concatenated source and target sentences are proportionally truncated to imitate a random start or break. This is governed by the computed partition values. By design, the truncation keeps most of the first sentence (lines 7, 8 for $S_1$, $T_1$: at most 30\% of the start of the first sentence is discarded) while discarding the bulk of the second sentence (lines 7, 8 for $S_2$, $T_2$: at most 30\% of the start of the second sentence is retained). This removes context from the first sentence $S_1$, adds context from the second sentence $S_2$, and combines them into a single training example. That new augmented sample will be pushed into the $L_{out}$ in line 9. After the algorithm passes through all sentences, it will return the list of augmented sentence pairs in line 11. Table \ref{tab:example_segmentation_da} shows how truncated sentences are formed to create an augmented sentence.

\begin{table}[H]
\tbl{Example of Sentence Boundary Augmentation applying for both Vietnamese and Bahnar.}
{\begin{tabular}{lp{5.6cm}p{5.6cm}} \toprule
Sentences & Source Sentence & Target Sentence\\ \midrule
$S_1$, $T_1$ & Phó Trưởng Ban thường trực: Ông Phan Trọng Hổ, Giám đốc Sở Nông nghiệp và Phát triển nông thôn & Phŏ Trương 'Ban thương trưk: 'Bok Phan Trong Hô, Giam đôk Sơ Nông nghiêp weng pơjing cham pơlĕi \\ \midrule
$S_2$, $T_2$ & Vì vậy, ngành y tế huyện, khuyến cáo người dân thận trọng trong việc sử dụng các loại nấm, tuyệt đối không được sử dụng các loại nấm lạ, để tránh bị ngộ độc & Yua noh, nganh y tê hŭn pơtho khan nă ma wă băt lơm tơdrong chă yuô rim loai mơu, pơgloh bi đĕi chă yuô rim loai mơu la sư hli ngô đôc\\ \midrule
$S_{out}$, $T_{out}$ & Ông Phan Trọng Hổ , Giám đốc Sở Nông nghiệp và Phát triển nông thôn . Vì vậy , ngành y tế huyện & 'Bok Phan Trong Hô , Giam đôk Sơ Nông nghiêp weng pơjing cham pơlĕi Yua noh , nganh y tê hŭn\\ \bottomrule
\end{tabular}}
\label{tab:example_segmentation_da}
\end{table}

In the original research, the authors may have solely focused on the algorithm. However, we aim to delve deeper into the effectiveness, especially concerning any changes in the parameter $p$, to ensure that the new augmented data does not diverge significantly, despite potential variations in the value of $p$.

\subsection{Pipeline}

\begin{figure}[H]
\centering
    \begin{tikzpicture}[
    SIR/.style={rectangle, draw=black!60, very thick,
    text centered,
    inner sep=0pt, 
    text width=5.5cm, 
    text height=5.5mm, 
    text depth=4.5mm
    },
    ]
    \node[SIR] (origin) {Original training data};
    \node[SIR] (methods) [below=of origin] {Applying DA methods};
    \node[SIR] (augmented) [below=of methods] {New augmented data};
    \node[SIR] (combining) [below=of augmented] {Augmented training set};
    \node[SIR] (training) [below=of combining] {Training};
    \node[SIR] (evaluation) [below=of training] {Evaluation};
    \node[SIR] (da_container) [right=of methods, minimum height=5cm] {};
    \node[SIR] (mtl) [below=of da_container.west, xshift=2.7cm, yshift=3.2cm, text width=4.7cm, text depth=4mm] { MTL DA };
    \node[SIR] (senbou)[below=0.1cm of mtl, text width=4.7cm, text depth=4mm] {Sentence Boundary};
    \node[SIR] (eda)[below=0.1cm of senbou, text width=4.7cm, text depth=4mm] {EDA};
    \node[SIR] (sem)[below=0.1cm of eda, text width=4.7cm, text depth=4mm] {Semantic Embedding};

    \draw[->, very thick] (origin.south) to node[right] {} (methods.north);
    \draw[->, very thick] (methods.south) to node[right] {} (augmented.north);
    \draw[->, very thick] (augmented.south) to node[right] {} (combining.north);
    \draw[->, very thick] (combining.south) to node[right] {} (training.north);
    \draw[->, very thick] (training.south) to node[right] {} (evaluation.north);
    \draw[->, very thick] (origin.west)  -- ++ (-1,0) |- node[pos=0.25,left] {Combining} (augmented.west);
    \draw[very thick, black, decoration={brace, raise=5pt, amplitude=3mm, mirror}, decorate] (da_container.north west) -- (da_container.south west) node [black, midway]{};
    
    \end{tikzpicture}
    \caption{General pipeline of augmenting, training and evaluating process}
    \label{fig:pipeline}
\end{figure}
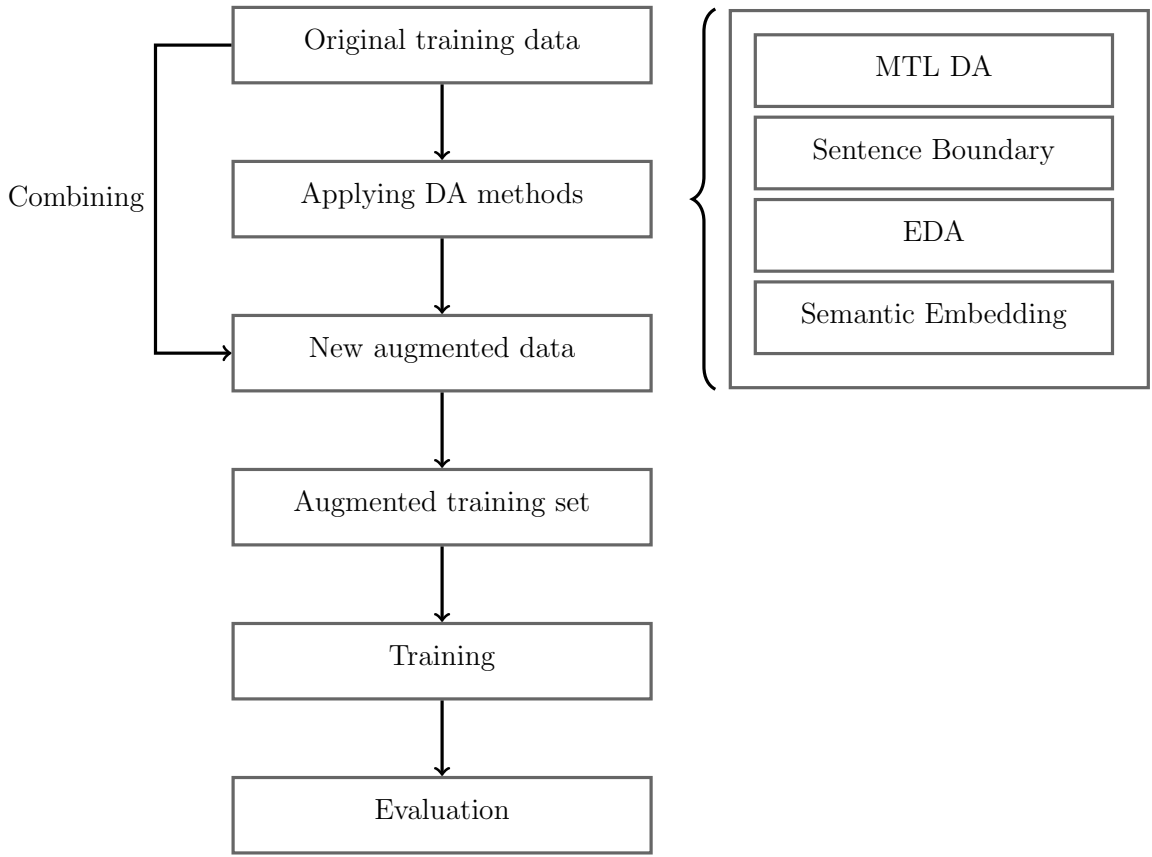

To investigate the effect of DA on translating performance, the training set will be augmented with a DA module that contains several DA methods, including MTL DA and sentence boundary augmentation. Besides, two more methods, EDA and semantic embedding, will also be experimented to analyze the effects. The new augmented training set is a combination of the original training data and the new augmented data. With each different DA approach, there will be a unique version of the augmented data set, and each of them will be passed through the same training process. In other words, this pipeline was designed to generate a number of new augmented datasets corresponding to each applied data augmentation method. After training, the results of the augmented data will be collected, compared, and evaluated to see the performance of each method's impact. Figure \ref{fig:pipeline} shows the general approach of how the researcher conducts this project.

\section{Experiments and discussion} \label{sec:experi}

\subsection{Dataset}

The dataset used for this project is the bilingual corpus of Vietnamese-Bahnar. The dataset contains a thousand lines of text in both Vietnamese and Bahnar. The dataset's sentences were initially collected from online news. The translated Bahnar version is made manually by the Bahnar collaborators based on the Bahnar Kriem dictionary. Later, the dataset was improved with more sentence pairs; these pairs were collected from the teaching textbook. These sentences are formal greetings, formal and informal conversations, narrative stories, and folktales written in Bahnar Kriem.

The dataset was divided into three sub-datasets: a training set, a test set, and a validation set, which were used for training, testing, and validating, respectively. Each sub-dataset contains two text files; the first text file is used for storing Vietnamese sentences (stored as .vi), and Bahnar sentences are stored in the other text file (stored as .ba). The total number of sentence pairs in each sub-set is shown in Table \ref{tab:origin_dataset}.

\begin{table}[H]
\tbl{Original Dataset}
{\begin{tabular}{cc} \toprule
 Type & \# of sentence pairs\\ \midrule
Training set & 16105 \\
    Test set & 1988 \\  
    Valid set & 1987 \\ \bottomrule
\end{tabular}}
\label{tab:origin_dataset}
\end{table}

\subsection{Evaluation metric} \label{subsec:bleu}

To evaluate experimental results, the main metric that is used for evaluating is \textit{BLEU}(Bilingual Evaluation Understudy) \cite{bleu}. BLEU is a metric for automatically evaluating machine-translated text. The BLEU score ranges from 0 to 1, representing how closely the machine-translated text resembles a set of excellent reference translations. Number 0 indicates little to no overlap between the machine-translated output and the reference translation (poor quality), whereas 1 indicates perfect overlap between the two translations (high quality). Sometimes, the decimal values are turned into a 0 to 100 scale so they can be read more easily; for example, 0.7 can be written as 70.

It has been demonstrated that BLEU scores and human evaluations of translation quality are highly correlated. It should be noted that even human translators do not get a score of 1.0. It is strongly advised against comparing BLEU scores across different corpora and languages. Even comparing BLEU scores within the same corpus but with varying numbers of reference translations can lead to highly misleading results. However, as a rough guideline, Table \ref{tab:bleu_inter} shows the interpretation of BLEU scores.

\begin{table}[H]
\tbl{Interpretation of BLEU scores \cite{google-a}}
{\begin{tabular}{ cc }  \toprule
    BLEU Score & Interpretation\\ \midrule
    < 0.10 & Almost useless\\ \midrule
    0.1 - 0.19 & Hard to get the gist\\ \midrule
    0.20 - 0.29 & The gist is clear,  but there are substantial grammatical errors present \\ \midrule
    0.30 - 0.40 & Understandable to good translations\\ \midrule
    0.40 - 0.50 & High quality translations\\ \midrule
    0.50 - 0.60 & Very high quality, adequate, and fluent translations\\ \midrule
    > 0.60 & Quality often better than human\\ \bottomrule
    \end{tabular}}
    \label{tab:bleu_inter}
\end{table}

\subsection{Experimental settings}

To experiment, \textit{Transformers} \footnote{visit https://github.com/huggingface/transformers} was chosen as the core framework. The framework compatibility of JAX, TensorFlow, and PyTorch is supported through transformers. With this flexibility, a model can be trained in one framework with a few lines of code and loaded for inference in another, depending on the stage of its existence. Google Colab \footnote{visit https://colab.research.google.com/} was chosen as an environment for training the model. 

The one-to-one word alignments required by \textit{replace} of MTL DA were obtained by using \textit{SimAlign} \footnote{https://github.com/cisnlp/simalign}. This alignment mechanism leverages multilingual word embeddings – both static and contextualized for word alignment \cite{jalili-sabet-etal-2020-simalign}. Multilingual embeddings are created from monolingual data only without relying on parallel data or dictionaries. One of the proposed default embedding models has been chosen for the embedding model used for the aligner, \textit{xlm-mlm-100-1280}.

During the training process, the author only tried to train and test the translation of Vietnamese-Bahnar using the model BARTpho. DA methods were studied and experimented with to see their effectiveness on translation performance. In the training process, $BARTpho_{syllable}$ was chosen as the pre-trained and baseline model, and the BLEU score is the primary metric to evaluate the best training model. All models were trained with the hyperparameters shown in Table \ref{tab:training_hyperparam}

\begin{table}[H]
\tbl{Training Hyperparameters}
{\begin{tabular}{p{3cm}p{1cm}} \toprule
Parameter & Value \\ \midrule
 Training batch size & 32\\
     Evaluation batch size & 32\\
     Epoch & 2\\
     Learning rate & 2e-5\\
     Weight decay & 0.01\\
     Evaluation steps & 1000\\
     Num beams & 5 \\
    Valid set & 1987 \\ \bottomrule
\end{tabular}}
\label{tab:training_hyperparam}
\end{table}

\subsection{Results and discussion} \label{section:result_discussion}

\subsubsection{General} \label{subsection:mtlda_exp}

To conduct the experiments, several augmented training sets were generated separately. With MTL DA and sentence boundary method, the size of generated data points was consistent. While MTL DA is twice as large as the original dataset size,  sentence boundary DA is only 0.67 times that of the origin (Shown in Table \ref{tab:result_training_set}). The augmented data size of the other two methods (EDA and semantic embedding) is followed by the exact value of MTL DA to evaluate fairly.

\begin{table}[H]
\tbl{Total sentence pairs of the baseline and augmented training sets}
    {\begin{tabular}{ cc } \toprule
    Method & \# of sentence pairs\\ \midrule
    Baseline & 16105 \\ \midrule
    MTL-DA & 32210 \\ \midrule
    Sentence boundary & 24157 \\ \midrule
    EDA & 32210 \\ \midrule
    Semantic embedding & 32210 \\ \bottomrule
    \end{tabular}}
    \label{tab:result_training_set}
\end{table}

Table \ref{tab:mtl_da} reports the translation performance, measured in terms of BLEU for prediction. In this experiment, hyperparameter $\alpha$ was chosen as 0.5 because if $\alpha$ is too small, nearly none of the transformations can apply to target sentences in the training set. So, augmented data will be very similar to original data, which either violates diversity criteria or harms future models in the out-domain fields. In contrast, if $\alpha$ is too large, the translating model of the in-domain can suffer from poor performance.

\begin{table}[H]
\tbl{BLEU scores of baseline and MTL DA approach, using different auxiliary tasks and their combinations}
    {\begin{tabular}{p{4cm}p{1.2cm}} \toprule
        Method & BLEU \\ \midrule
        Baseline & 29.89 \\ \midrule
        Swap & 35.22 \\ \midrule
        \textbf{Token} & \textbf{38.48} \\ \midrule
        Source & 2.38 \\ \midrule
        Reverse & 31.14 \\ \midrule
        \textbf{Replace} & \textbf{38.01} \\ \midrule
        Replace+swap & 37.91 \\ \midrule
        Replace+token & 38.35 \\ \midrule
        \textbf{Token+swap} & \textbf{40.64} \\ \midrule
        Replace+token+swap & 37.03 \\ \bottomrule
    \end{tabular}}
    \label{tab:mtl_da}
\end{table}

First of all, the baseline is the evaluation results when training and testing without any augmentation method applied. The results show that the MTL DA approach consistently outperforms the baseline system except for method \textit{source}. Generally, the auxiliary tasks \textit{token} and \textit{replace} are the best-performing ones. \textit{swap} and \textit{reverse} may give a lower performance result than these two methods above, which suggests that abnormal word order could negatively influence the main task, but it still produces a promising result. While all five methods have shown improved results, \textit{source} has its worst performance, which indicates that the translation task could be adversely affected by introducing a completely different vocabulary in the target.

Interestingly, using each two of the three best auxiliary tasks together further improves the performance, achieving well-performed results in all translation tasks with BLEU scores between 8.02(replace+swap) and 10.75 (token+swap) points over the baseline. The combination of \textbf{token+swap} gives the best performance. Although the researcher has combined all three best methods, the results still cannot outperform the combination of token and swap. In general, all combinations have led to an enhancement in the BLEU score. This suggests that different auxiliary tasks have unique effects on the encoder and show a form of mutual support.

In \textit{sentence boundary augmentation}, DA performs based on truncated sentence combination; this combination depends on hyperparameter $p$. Therefore, the researcher has examined different values of $p$ on the results (shown in Table \ref{tab:sentence_boundary_da}). Surprisingly, the values of $p$ do not affect so much on translation results. The BLEU scores of different $p$ values are slightly different, although, in this specific project, $p=0.7$ has shown the most well-performed results compared with others. This method has shown a significant advance compared to the baseline system.

\begin{table}[H]
\tbl{BLEU scores obtained using different $p$ in sentence boundary augmentation approach}
    {\begin{tabular}{p{4cm}p{1.2cm}} \toprule
        $p$ & BLEU \\ \midrule
        {$p=0.1$}  & 40.34  \\ \midrule
        {$p=0.3$}  & 39.86  \\ \midrule
        {$p=0.5$}  & 40.47  \\ \midrule
        {$p=0.7$}  & \textbf{41.33} \\ \midrule
        {$p=0.9$}  & 40.86 \\ \bottomrule
    \end{tabular}}
    \label{tab:sentence_boundary_da}
\end{table}

Considering the same training configuration, with the less generated sentence pairs, the best result of \textit{sentence boundary augmentation} can be on pair with any auxiliary task from MTL DA and their combinations. It indicates that \textit{sentence boundary augmentation} can perform well in the context of low-resource translation at least. It can also utilize limited resources to produce a smaller augmented training set. For low-resource machine translation augmentation, the nosing-based method applied at the phrase or sentence level has demonstrated superior performance to the nosing-based method applied at the word level.

The researcher chooses hyperparameter $\alpha = 0.5$ for EDA and semantic embedding. Both methods were applied on the target site to serve the same purpose of MTL DA: to strengthen the encoder and make a word dependent less on the prefix word. Four methods can overcome the baseline, but EDA has fallen behind the other three methods (shown in Table \ref{tab:all_da}). This issue happens due to such reasons: random insertion and random replacement need to use the external resource from the dictionary, which may provide some new words to the vocabulary; random swap and random deletion could accidentally change the word; these four methods combined together could create a negative effect. Therefore, EDA, initially designed for text classification, can not be utilized in the context of NMT. Semantic embedding may have good performance but still cannot compare with MTL DA combination and sentence boundary, which can prove that word replacement is a possible and promising solution, but it needs to have reasonable strategies to decide how the method should be used, such as augmented language site, hyperparameter, utilizing language model, strategy in choosing words to replace, etc.

\begin{table}[H]
\tbl{BLEU scores obtained with baseline, MTL DA, sentence boundary, EDA, and semantic embedding}
    {\begin{tabular}{p{4cm}p{1.2cm}} \toprule 
        Method & BLEU \\ \midrule
        Baseline & 29.89 \\ \midrule
        EDA & 36.37 \\ \midrule
        Semantic embedding & 39.20\\ \midrule
        Token+swap & 40.64 \\ \midrule
        Sentence boundary & 41.33 \\ \bottomrule
    \end{tabular}}
    \label{tab:all_da}
\end{table}

\subsubsection{Sentence orientation}
Section \ref{sec:vb_tran} has mentioned several issues that might happen during the translating process, such as collocation issues, word-by-word translating issues, etc. Therefore, this project will also focus on improving these specific cases. However, it is impossible to cover all the mentioned issues in the whole testing set; only a set of sentences is chosen to investigate. Based on suggestions from Table \ref{tab:bleu_inter}, the researcher chose the translated sentences with a BLEU score range from 0.2 to 0.4. Because it is complicated to detect translating issues with a low BLEU score sentence; on the other hand, a sentence with a high BLEU score is good, making it hard to exploit issues. Based on the given criteria, there are only 222 satisfied sentences from the test set. Table \ref{tab:test_issue} shows each issue and its total sentences.

\begin{table}[H]
\tbl{Translating issues of chosen sentences in test set}
    {\begin{tabular}{p{4cm}p{2cm}} \toprule 
        Issue & \# sentences \\ \midrule
        Collocation & 34 \\ \midrule
        Word-by-Word & 36  \\ \midrule
        Number ambiguity & 102\\ \midrule
        Unknown & 50  \\ \bottomrule
    \end{tabular}}
    \label{tab:test_issue}
\end{table}

Collocation and word-by-word have been mentioned before, and the leftover is "number ambiguity" and "unknown". "Number ambiguity" refers to the issue that numbers are written as text in the test set, but their prediction is in number. Because using the BLEU score, which relies heavily on tokens, the BLEU score of these sentences is really low, although other parts are translated well. "Unknown" can be understood as unstable in translating a long sentence that is hard to identify which issue occurs within. Therefore, in this section, experiments will only focus on two issues: Collocation and word-by-word.

\begin{table}[H]
\tbl{BLEU scores of Collocation and word-by-word with baseline and other DA methods}
    {\begin{tabular}{ccc} \toprule
        Method & Collocation & word-by-word \\ \midrule
        baseline & 1.52 & 6.61 \\ \midrule
        token+swap & 4.64 & 9.19 \\ \midrule
        sentence boundary & 3.87 & 11.08\\ \bottomrule
    \end{tabular}}
    \label{tab:sentence_predict_bleu}
\end{table}

Table \ref{tab:sentence_predict_bleu} represents all the experiments of baseline and DA augmented models to two issues, "collocation" and "word-by-word". In this experiment, the researcher only focuses on MTL DA(token+swap) and sentence boundary augmentation because both have proved efficient compared to baseline and other methods. In general, the predicted BLEU scores of "collocation" are always smaller than "word-by-word" in both baseline and augmentation contexts, which indicates that the collocation issue is more complicated to handle than the other one. Overall, two augmentation methods have significantly improved the above linguistic issues. 

For collocation, the MTL DA-based method - "token+swap" grows over three times higher than the baseline, while "sentence boundary" is 2.5 times higher when compared with the baseline. On the other hand, for "word-by-word", "sentence boundary" reaches almost 68\% better performance when compared with the baseline; however, the "token+swap" method can only achieve nearly 40\% higher than the baseline. Interestingly, each context has its own dominant DA method. For collocation, the effective DA method was "token+swap", while "sentence boundary" was the method that gave the best support to "word-by-word". There is a reasonable explanation for both cases. With word-by-word, the issue comes from the fraction translations, which keep the same original words; however, "sentence boundary" can augment the data and produce more stable sentence-level truncation simultaneously. Therefore, the augmented models can learn from the new set of words. Noticing that "sentence boundary" in "collocation" has a lower performance than MTL DA, this issue happens because the focused collocations in the experiment are usually allocated at the beginning or the end of the sentence, which might be truncated due to the default algorithm of sentence boundary (see in Algorithm \ref{alg:sentence_boundary_augmentation}).

\section{Conclusion} \label{sec:conclu}
The main point of this paper is data augmentation for neural machine translation, especially data augmentation methods for translating low-resource languages, which is translating Vietnamese to Bahnar in this context. Several types of research have been studied; the DA methods have been addressed from word replacing, nosing-based method, to back-translation. Besides, the attributes of the Bahnar language have been studied to explore more linguistic features for conducting this project. In this project, two approaches have been suggested and effectively implemented for both scenarios: general sentences and oriented sentences.

Firstly, \textit{Multi-task Learning Data Augmentation}, known as MTL DA, differs from conventional approaches that focus on expanding the support of empirical data distribution by generating new samples that align with this distribution. These techniques produce new sentence pairs by applying radical transformations synthetically. The translation of augmented sentences introduces fresh contexts during training. When translating augmented sentences during training, the system learns to rely more on the encoder because the target prefix alone is insufficient to predict the next word accurately. As a result, the encoder is reinforced, and the system is compelled to rely more on it. Experiments were carried out on five low-resource translation tasks, four of which have improved over the baseline system. The other one (\textit{source}) represents a harmful affection when adding a whole new vocabulary on the target side.

Secondly, \textit{Sentence Boundary Augmentation}, which explores the noising-base method (\textit{swap}) beyond word level by applying sentence level for augmenting. This is also a significant point when a method with one purpose (overcoming wrong segmentation for translation in the ASR system) can apply for another purpose (low-resource NMT), and it even makes an impressive performance result (improving the average 10 BLEU score compared with the baseline). This interference presents an opportunity for exploring and applying methods from other aspects to enhance low-resource NMT. 

Both of these approaches are independent of the specific architecture of the NMT model. Furthermore, they do not necessitate complex preprocessing steps, training of additional systems, or acquiring additional data beyond the existing training parallel corpora. However, most of the proposed operations are implemented in the target language; only two methods are applied on both sides. Furthermore, it could be combined with existing DA methods, such as back-translation, especially those that operate on the source side \cite{wang-etal-2018-switchout, gao-etal-2019-soft}.

Besides, further improvements could be achieved by implementing more sophisticated approaches to multi-task learning, such as changing the proportion of data for the different tasks and evaluating different ways of parameter sharing between the different tasks. Moreover, to evaluate the validity of these methods, different low-resource datasets (e.g. different dialects, different low-resource languages) should be trained and tested.

\section*{Competing interests}
The author(s) declare that they have no competing interests.

\section*{Acknowledgment(s)}
This research is funded by the Ministry of Science and Technology (MOST) within the framework of the Program "Supporting research, development and technology application of Industry 4.0" KC-4.0/19-25 – Project “Development of a Vietnamese- Bahnaric machine translation and Bahnaric text-to-speech system (all dialects)” - KC-4.0-29/19-25.


\bibliographystyle{vancouver}
\bibliography{ref}

\end{document}